\documentclass{article}
\usepackage{spconf,amsmath,graphicx}

\title{DenseNet for Dense Flow}
\name{Yi Zhu and Shawn Newsam}

\address{University of California, Merced \\
	5200 N Lake Rd, Merced, CA, US  \\
	\{yzhu25, snewsam\}@ucmerced.edu }
\begin{document}
%
\maketitle
\begin{abstract}
Classical approaches for estimating optical flow have achieved rapid progress in the last decade. However, most of them are too slow to be applied in real-time video analysis. 
Due to the great success of deep learning, recent work has focused on using CNNs to solve such dense prediction problems.
In this paper, we investigate a new deep architecture, Densely Connected Convolutional Networks (DenseNet), to learn optical flow. This specific architecture is ideal for the problem at hand as it provides shortcut connections throughout the network, which leads to implicit deep supervision. 
We extend current DenseNet to a fully convolutional network to learn motion estimation in an unsupervised manner. 
Evaluation results on three standard benchmarks demonstrate that DenseNet is a better fit than other widely adopted CNN architectures for optical flow estimation.
\end{abstract}
\begin{keywords}
Optical flow estimation, Unsupervised learning, Convolutional neural network
\end{keywords}
\section{Introduction}
\label{sec:intro}
Convolutional Neural Networks (CNNs), due to their immense learning capacity and superior efficiency, have advanced a variety of computer vision tasks, including optical flow prediction. 
Recent work \cite{flownet,sceneflow2016} built large-scale synthetic datasets to train a supervised CNN and show that networks trained on such unrealistic data still generalize very well to existing datasets such as Sintel \cite{mpi_sintel} and KITTI \cite{Geiger2012CVPR}. Other works \cite{AhmadiICIP2016,jasonUnsup2016,guided_zhu_2017} have designed new objectives such as image reconstruction loss to guide the network learning in an unsupervised way for motion estimation. Though \cite{flownet,sceneflow2016,AhmadiICIP2016,jasonUnsup2016} are totally different approaches, they all use variants of one architecture, the ``FlowNet Simple'' network \cite{flownet}.

FlowNetS is a conventional CNN architecture, consisting of a contracting part and an expanding part. Given adjacent frames as input, the contracting part uses a series of convolutional layers to extract high level semantic features, while the expanding part tries to predict the optical flow at the original image resolution by successive deconvolutions. In between, it uses skip connections \cite{fcn_cvpr15} to provide fine image details from lower layer feature maps. This generic pipeline, \textit{contract, expand, skip connections}, is widely adopted for per-pixel prediction problems, such as semantic segmentation \cite{tiramisu_16}, depth estimation \cite{monodepthlr2016}, video coloring \cite{V2V_CVPR16}, etc. 

However, skip connections are a simple strategy for combining coarse semantic features and fine image details; they are not involved in the learning process. 
What we desire is to keep the high frequency image details until the end of the network in order to provide implicit deep supervision. Simply put, we want to ensure maximum information flow between layers in the network.

DenseNet \cite{densenet_16}, a recently proposed CNN architecture, has an interesting connectivity pattern: each layer is connected to all the others within a dense block. In this case, all layers can access feature maps from their preceding layers which encourages heavy feature reuse. As a direct consequence, the model is more compact and less prone to overfitting. Besides, each individual layer receives direct supervision from the loss function through the shortcut paths, which provides implicit deep supervision. All these good properties make DenseNet a natural fit for per-pixel prediction problems. There is a concurrent work using DenseNet for semantic segmentation \cite{tiramisu_16}, which achieves state-of-the-art performance without either pretraining or additional post-processing. However, estimating optical flow is different from semantic segmentation. We will illustrate the differences in Section \ref{sec:experiments}. 

In this paper, we propose to use DenseNet for optical flow prediction.
Our contributions are two-fold. 
First, we extend current DenseNet to a fully convolutional network. Our model is totally unsupervised, and achieves performance close to supervised approaches. 
Second, we empirically show that replacing convolutions with dense blocks in the expanding part yields better performance. 

\begin{figure*}[t]
	\centering
	\includegraphics[width=1.0\linewidth,trim=0 0 0 0,clip]{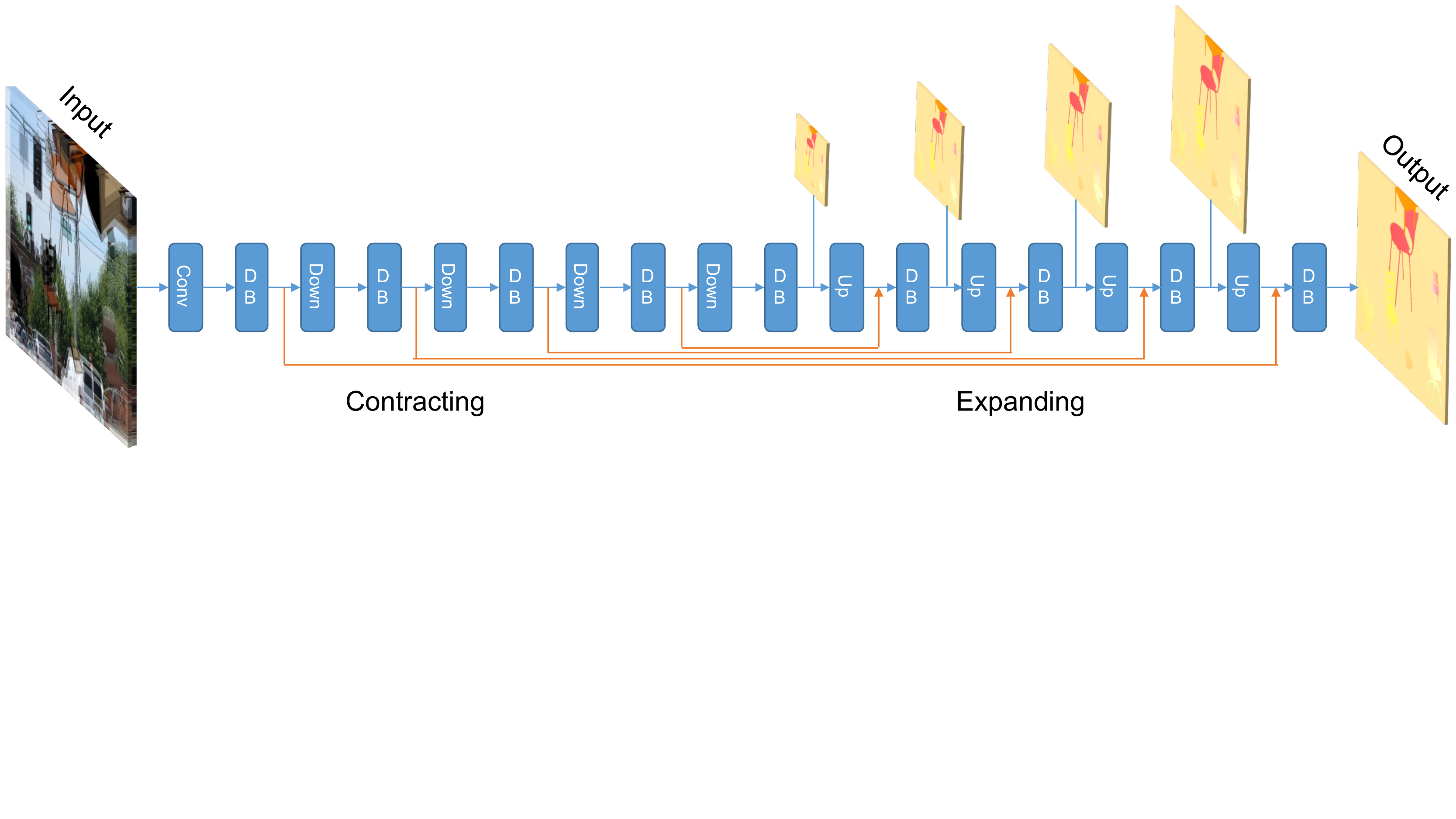}
	\vspace{-32ex}
	\caption{An overview of our unsupervised learning framework based on dense blocks (DB). ``Down'' is the transition down layer, and ``Up'' is the transition up layer. The orange colored arrows indicate the skip connections. See more details in Section \ref{sec:densenet_fc}. }
	\label{fig:overview}
\end{figure*}

\section{Method}
\label{sec:method}
Given adjacent frames, previous $I_{1}$ and next $I_{2}$, our goal is to learn a model that can predict per-pixel motion field $(U, V)$ between the two images. $U$ and $V$ are the horizontal and vertical displacement. In this section, we first review the DenseNet architecture, and then outline our unsupervised learning framework based on a fully convolutional DenseNet.

\subsection{DenseNet Review}
\label{sec:densenet_review}
Traditional CNNs, such as FlowNetS, calculate the output of the $l^{th}$ layer by applying a nonlinear transformation $H$ to the previous layer's output $x_{l-1}$,
\begin{equation}
x_{l} = H_{l} (x_{l-1}).
\label{eq:standard_cnn}
\end{equation}
Through consecutive convolution and pooling, the network achieves spatial invariance and obtains coarse semantic features in the top layers. However, fine image details tend to disappear in the very top of the network. 

To improve information flow between layers, DenseNet \cite{densenet_16} provides a simple connectivity pattern: the $l^{th}$ layer receives the feature maps of all preceding layers as inputs:
\begin{equation}
x_{l} = H_{l} ([x_{0}, x_{1}, ... ,x_{l-1}])
\label{eq:densenet}
\end{equation}
where $[x_{0}, x_{1}, ... ,x_{l-1}]$ is a single tensor constructed by concatenation of the previous layers' output feature maps. In this manner, even the last layer can access the input information of the first layer. And all layers receive direct supervision from the loss function through the shortcut connections. 
$H_{l}(\cdot)$ is a composite function of four consecutive operations, batch normalization (BN), leaky rectified linear units (LReLU), a $3\times3$ convolution and dropout. We denote such composite function as one layer. 

In our experiments, the DenseNet in the contracting part has four dense blocks, each of which has four layers. Between the dense blocks, there are transition down layers consisting of a $1\times1$ convolution followed by a $2\times2$ max pooling. We compare DenseNet with three other popular architectures, namely FlowNetS \cite{flownet}, VGG16\cite{vgg_iclr15} and ResNet18 \cite{resnet_cvpr16} in Section \ref{sec:discussion}. 

\subsection{Fully Convolutional DenseNet}
\label{sec:densenet_fc}
Classical expanding uses series of convolutions, deconvolutions, and skip connections to recover the spatial resolution in order to get the per-pixel prediction results. Due to the good properties of DenseNet, we propose to replace the convolutions with dense blocks during expanding as well.

However, if we follow the same dense connectivity pattern, the number of feature maps after each dense block will keep increasing. Considering that the resolution of the feature maps also increases during expanding, the computational cost will be intractable for current GPUs. 
Thus, for a dense block in the expanding part, we do not concatenate the input to its final output. For example, if the input has $k_{0}$ channels, the output of an $L$ layer dense block will have $L k$ feature maps. k is the growth rate of a DenseNet, defining the number of feature maps each layer produces. Note that dense blocks in the contracting part will output $k_{0} + Lk$ feature maps.

For symmetry, we also introduce four dense blocks in the expanding part, each of which has four layers. The bottom layer feature maps at the same resolution are concatenated through skip connections. Between the dense blocks, there are transition up layers composed of two $3\times3$ deconvolutions with a stride of $2$. One is for upsampling the estimated optical flow, and the other is for upsampling the feature maps. 

\subsection{Unsupervised Motion Estimation}
\label{sec:densenet_unsup}
Supervised approaches adopt synthetic datasets for CNNs to learn optical flow prediction. However, synthetic motions/scenes are quite different from real world ones, thus limiting the generalizability of the learned model. Besides, even constructing synthetic datasets requires a lot of manual effort \cite{mpi_sintel}. Hence, unsupervised learning is an ideal option for the naturally ill-conditioned motion estimation problem.

Recall that the unsupervised approach \cite{jasonUnsup2016} treats the optical flow estimation as an image reconstruction problem. The intuition is that if we can use the predicted flow and the next frame to reconstruct the previous frame, our network is learning useful representations about the underlying motions. 
To be specific, we denote the reconstructed previous frame as $I_{1}^{\prime}$. The goal is to minimize the photometric error between the previous frame $I_{1}$ and the inverse warped next frame $I_{1}^{\prime}$: 
\begin{equation}
L_{\text{reconst}} = \frac{1}{N} \sum_{i, j}^{N} \rho ( I_{1}(i, j) - I_{1}^{\prime}(i,j) ).
\label{eq:reconstruction_loss}
\end{equation}
Here $I_{1}^{\prime}(i,j) = I_{2}(i+U_{i,j}, j+V_{i,j})$. N is the total number of pixels. The inverse warp is done by using spatial transformer modules \cite{stn_nips15} inside the CNN. We use a robust convex error function, the generalized Charbonnier penalty $\rho(x) = (x^{2} + \epsilon^{2})^{\alpha}$, to reduce the influence of outliers. This reconstruction loss is similar to the brightness constancy objective in classical variational formulations.

An overview of our unsupervised learning framework based on DenseNet is illustrated in Fig. \ref{fig:overview}. Our network has a total of $53$ layers with a growth rate of $12$. But due to the parameter efficiency of dense connectivity, our model only has $2$M parameters, while FlowNetS has $38$M.

\section{Experiments}
\label{sec:experiments}

\subsection{Datasets}
\label{sec:datasets}
\textbf{Flying Chairs} \cite{flownet} is a synthetic dataset designed specifically for training CNNs to estimate optical flow. It is created by applying affine transformations to real images and synthetically rendered chairs. The dataset contains 22,872 image pairs: 22,232 training and 640 test samples according to the standard evaluation split. 

\noindent \textbf{MPI Sintel} \cite{mpi_sintel} is also a synthetic dataset derived from a short open source animated 3D movie. There are 1,628 frames, 1,064 for training and 564 for testing. 
In this work, we only report performance on its final pass because it contains sufficiently realistic scenes including natural image degradations.

\noindent \textbf{KITTI Optical Flow 2012} \cite{Geiger2012CVPR} is a real world dataset collected from a driving platform. It consists of 194 training image pairs and 195 test pairs with sparse ground truth flow. We report the average endpoint error (EPE) on the entire image. 

Since the dataset size of Sintel/KITTI is relatively small, we first pretrain our network on Chairs, and then fine tune it to report the performance. Note that the fine tuning here is also unsupervised-we are not using ground truth flow from Sintel/KITTI.

\subsection{Implementation}
\label{sec:implementation}

During unsupervised training, we calculate the reconstruction loss for each expansion. There are $5$ expansions in our network, resulting in $5$ losses. We use the same loss weights as in \cite{jasonUnsup2016}. The generalized Charbonnier parameter $\alpha$ is $0.25$ and $\epsilon$ is $0.001$. The models are trained using Adam optimization with default parameters, $\beta_{1}=0.9$ and $\beta_{2}=0.999$. The initial learning rate is set to $10^{-5}$, and then divided by half every $100$k. We end our training at $600$k iterations.
We apply the same data augmentations as in \cite{jasonUnsup2016} to prevent overfitting.

\begin{figure*}[t]
	\centering
	\includegraphics[width=1.0\linewidth,trim=0 0 0 0,clip]{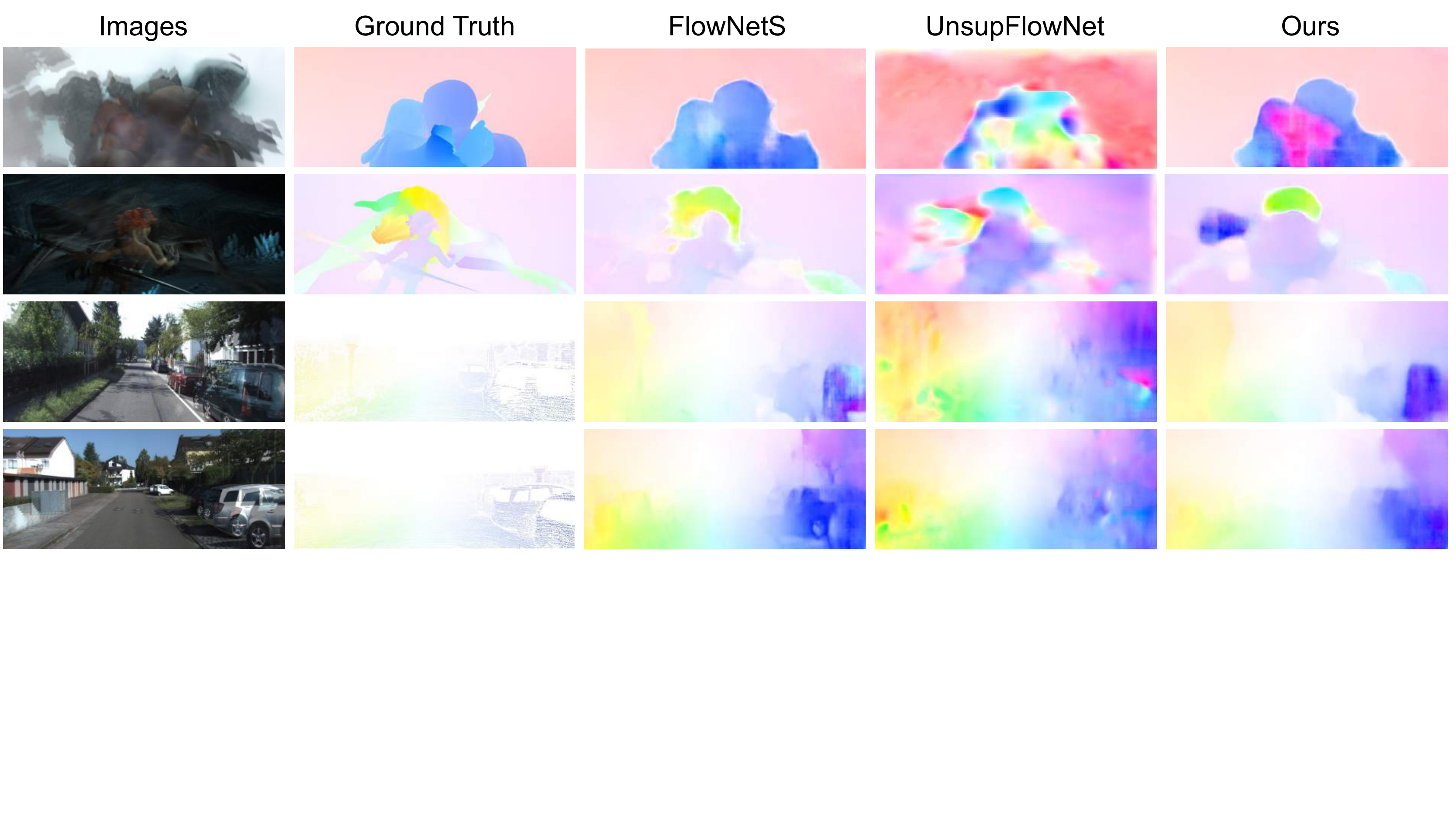}
	\vspace{-24ex}
	\caption{Visual examples of predicted optical flow from different methods. Top two are from Sintel, and bottom two from KITTI. }
	\label{fig:comparison}
	\vspace{-2ex}
\end{figure*}

\begin{table}
	\begin{center}
		\resizebox{\columnwidth}{!}{%
			\begin{tabular}{ | c | c  | c | c |}
				\hline
				Method																								&    Chairs    &    Sintel    & KITTI \\
				\hline		
				UnsupFlowNet \cite{jasonUnsup2016}													&   $5.30$ 	&  $11.19$  & $12.4$\\
				VGG16	\cite{vgg_iclr15}															 &   $5.47$ 		 & $11.35$  & $12.7$\\
				ResNet18 	\cite{resnet_cvpr16}															   &   $5.22$ 		 & $10.98$  & $12.3$\\
				DenseNet	\cite{densenet_16}													&   $5.01$ 		 & $10.66$  & $12.1$\\
				\hline
				DenseNet + Dense Upsampling	 												&   $\mathbf{4.73}$ 		 & $\mathbf{10.07}$  & $\mathbf{11.6}$\\
				DenseNet + Dense Upsampling (Deeper)									&   $6.65$ 		 & $13.46$  & $14.0$\\
				\hline
			\end{tabular}
		}
		\vspace{0ex}
		\caption{Optical flow estimation results on the test set of Chairs, Sintel and KITTI. All performances are reported using average EPE, lower is better.  Top: Comparison of different architectures with classical upsampling. Bottom: Our proposed DenseNet with dense block upsampling. \label{tab:result1}}
		\vspace{-5ex}
	\end{center}
\end{table} 

\vspace{-2ex}
\subsection{Results and Discussion}
\label{sec:discussion}
We have three observations given the results in Table \ref{tab:result1}.

\noindent \textbf{Observation $\mathbf{1}$}: As shown in the top section of Table \ref{tab:result1}, all four popular architectures perform reasonably well on optical flow prediction. The reason why VGG16 performs the worst is that multiple pooling layers may lose the image details. On the contrary, ResNet18 only has one pooling layer in the beginning, so it performs better than both VGG16 and FlowNetS. Interestingly, DenseNet also has multiple pooling layers, but due to dense connectivity, we don't lose fine appearance information. Thus, as expected, DenseNet performs the best with the least number of parameters.

Inspired by success of using deeper models, we also implement a network with five dense blocks in both the contracting and expanding parts, where each block has ten layers. However, as shown in the last row of Table \ref{tab:result1}, the performance is much worse due to overfitting. This may indicate that optical flow is a low-level vision problem, that doesn't need a substantially deeper network to achieve better performance. 

\noindent \textbf{Observation $\mathbf{2}$}: Using dense blocks during expanding is beneficial. In Table \ref{tab:result1}, DenseNet with dense upsampling achieves better performance on all three benchmarks than DenseNet with classical upsampling, especially on Sintel. As Sintel has much more complex context than Chairs and KITTI, it may benefit more from the implicit deep supervision.  This confirms that using dense blocks instead of a single convolution can maintain more information during the expanding process, which leads to better flow estimates.

\noindent \textbf{Observation $\mathbf{3}$}: One of the advantages of DenseNet is that it is less prone to overfitting. The authors in \cite{densenet_16} have shown that it can perform well even when there is no data augmentation compared to other network architectures. We investigate this by directly training from scratch on Sintel, without pretraining using Chairs. We built the training dataset using image pairs from both the final and clean passes of Sintel. When we use the same implementation and training strategies, the flow estimation performance is $10.3$, which is very close to $10.07$. One possible reason for such robustness is because of the model compactness and implicit deep supervision provided by DenseNet. This is ideal for optical flow estimation since most benchmarks have limited training data.

\subsection{Comparison to State-of-the-Art}
In this section, we compare our proposed method to recent state-of-the-art approaches. We only consider approaches that are fast because optical flow is often used in time sensitive applications. 
We evaluated all CNN-based approaches on a workstation with an Intel Core I7 with 4.00GHz and an Nvidia Titan X GPU. For classical approaches, we use their reported runtime. 

As shown in Table \ref{tab:result2}, although unsupervised learning still lags behind supervised approaches \cite{flownet}, our network based on fully convolutional DenseNet shortens the performance gap and achieves lower EPE on the three standard benchmarks than the state-of-the-art unsupervised approach \cite{jasonUnsup2016}. Compared to \cite{AhmadiICIP2016}, we get a higher EPE on Sintel because they use a variational refinement technique.

We show some visual examples in Figure \ref{fig:comparison}. We can see that supervised FlowNetS can estimate optical flow close to the ground truth, while UnsupFlowNet struggles to maintain fine image details and generates very noisy flow estimation. Due to the dense connectivity pattern, our proposed method can produce much smoother flow than UnsupFlowNet, and recover the high frequency image details, such as human boundaries and car shapes. 

Therefore, we demonstrate that DenseNet is a better fit for dense optical flow prediction, both quantitatively and qualitatively. 
However, by exploring different network architectures, we found that existing networks perform similarly on predicting optical flow. We may need to design new operators like the correlation layer \cite{flownet} or novel architectures \cite{spynet_16,flownet2} to learn motions between adjacent frames in future work. The model should handle both large and small displacement, as well as fine motion boundaries. Another concern of this work is that DenseNet has a large memory bandwidth which may limit its potential applications like action recognition \cite{depth2action,act_det_zhu_wacv17,hidden_zhu_17}.

\begin{table}
	\begin{center}
		\resizebox{0.9\columnwidth}{!}{%
			\begin{tabular}{ | c | c | c | c | c | }
				\hline
				Method										&    Chairs      &    Sintel    & KITTI  & Runtime  \\
				\hline
				EPPM 	\cite{EPPM_cvpr14}							&   $-$ 		 & $8.38$    & $9.2$    &  $0.25$\\	
				PCA-Flow 	\cite{pca_flow_cvpr15}						&   $-$ 		 & $8.65$    & $6.2$    &  $0.19^{\ast}$\\	
				DIS-Fast \cite{dis_fast_eccv16}							&   $-$ 		 & $10.13$   & $14.4$    &  $0.02^{\ast}$\\	
				\hline
				FlowNetS \cite{flownet} 							&   $2.71$ 		 & $8.43$   & $9.1$    &  $0.06$\\	
				USCNN \cite{AhmadiICIP2016}								&   $-$ 	 & $8.88$   & $-$    &  $-$\\	
				UnsupFlowNet \cite{jasonUnsup2016}								&   $5.30$ 	 & $11.19$   & $12.4$    &  $0.06$\\
				Ours		&   				$4.73$ 	 & $10.07$   & $11.6$      &  $0.13$\\ 
				\hline
			\end{tabular}
		}
		\vspace{0ex}
		\caption{State-of-the-art comparison. Runtime is reported in seconds per frame.  Top: Classical approaches. Bottom: CNN-based approaches. $^{\ast}$ indicates the algorithm is evaluated using CPU, while the rest are on GPU. \label{tab:result2}}
		\vspace{-6ex}
	\end{center}
\end{table}

\section{Conclusion}
\label{sec:conclusion}
In this paper, we extend the current DenseNet architecture to a fully convolutional network, and use image reconstruction loss as guidance to learn motion estimation. Due to the dense connectivity pattern, our proposed method achieves better flow accuracy than the previous best unsupervised approach \cite{jasonUnsup2016}, and shortens the performance gap with supervised ones. Besides, our model is totally unsupervised. Thus we can experiment with large-scale video corpora in future work, to learn non-rigid real world motion patterns. Through comparison of popular CNN architectures, we found that it is important to design novel operators or networks for optical flow estimation instead of relying on existing architectures for image classification.

\section{Acknowledgements}
This work was funded by a National Science Foundation CAREER grant, $\#$IIS-1150115. We gratefully acknowledge the support of NVIDIA Corporation through the donation of the Titan X GPU used in this work.


\bibliographystyle{IEEEbib}
\bibliography{strings,refs}

\end{document}